%% file: bmvc_review.tex
\documentclass{bmvc2k}

\title{DiffusedWrinkles: A Diffusion-Based \\Model for Data-Driven Garment Animation}

\addauthor{Raquel Vidaurre}{raquel.vidaurre@urjc.es}{1}
\addauthor{Elena Garcés}{elena.garces@urjc.es}{1}
\addauthor{Dan Casas}{dan.casas@urjc.es}{1}

\addinstitution{
 Universidad Rey Juan Carlos\\
 Madrid, Spain
}

\runninghead{Vidaurre, Garcés, Casas}{DiffusedWrinkles}

\def\etal{\emph{et al}\bmvaOneDot}

\usepackage{upgreek}
\usepackage{graphicx}
\usepackage{booktabs}
\usepackage{tikz}
\usepackage{amsfonts}
\usepackage{wrapfig}
\newcommand{\vect}[1]{\mathbf{#1}}

\begin{document}

\maketitle

\vspace{-0.40cm}
\begin{abstract}
We present a data-driven method for learning to generate animations of 3D garments using a 2D image diffusion model. In contrast to existing methods, typically based on fully connected networks, graph neural networks, or generative adversarial networks, which have difficulties to cope with parametric garments with fine wrinkle detail, our approach is able to synthesize high-quality 3D animations for a wide variety of garments and body shapes, while being agnostic to the garment mesh topology. Our key idea is to represent 3D garment deformations as a 2D layout-consistent texture that encodes 3D offsets with respect to a parametric garment template. Using this representation, we encode a large dataset of garments simulated in various motions and shapes and train a novel conditional diffusion model that is able to synthesize high-quality pose-shape-and-design dependent 3D garment deformations. Since our model is generative, we can synthesize various plausible deformations for a given target pose, shape, and design. Additionally, we show that we can further condition our model using an existing garment state, which enables the generation of temporally coherent sequences.

\end{abstract}

\begin{wrapfigure}[14]{r}{0.52\linewidth}
\vspace{-0.9cm}
\centering
  \resizebox{\linewidth}{!}{%
\begin{tikzpicture}
    \node (img) at (0.5,0.5) {\includegraphics[trim = {800, 0 0 0}, clip, width=\linewidth]{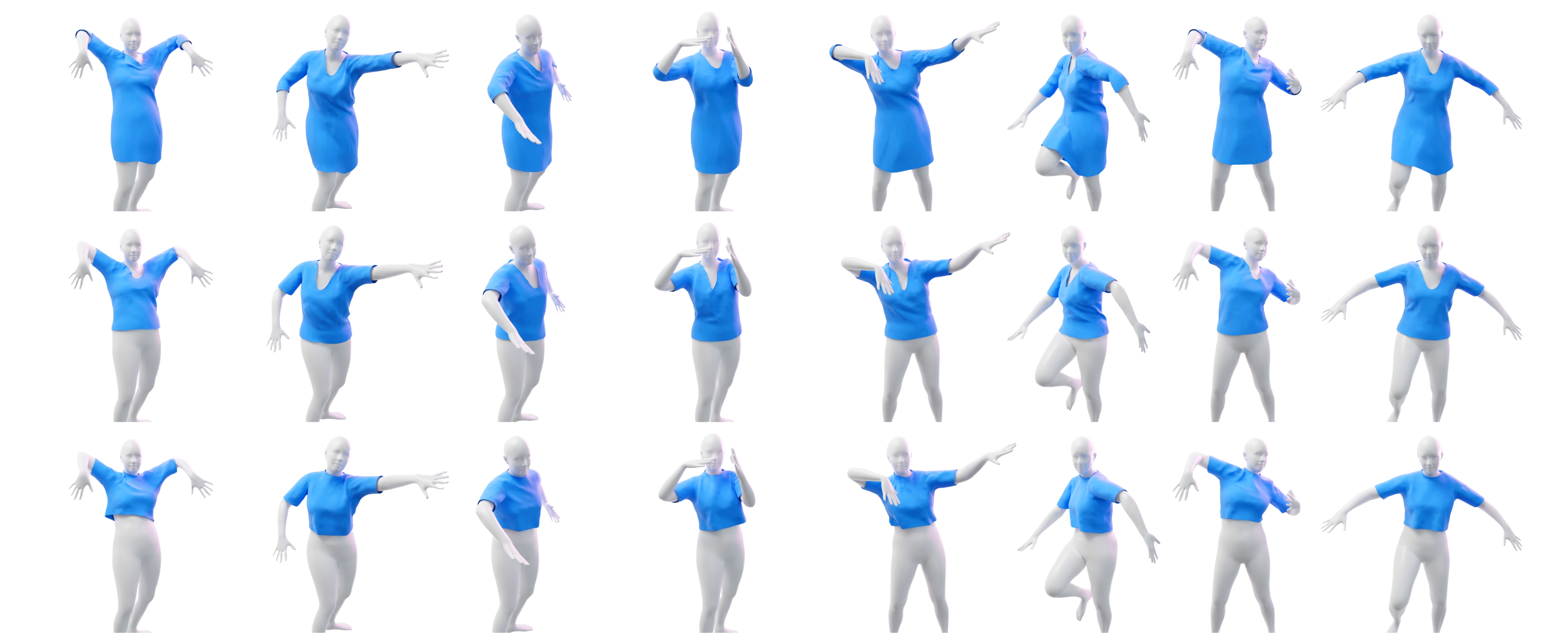}};
    \node (input)[left, align=center, xshift = 5pt, yshift = 70pt, rotate=90] at (img.west){Design A};
    \node (input)[left, align=center, xshift = 5pt, yshift = 20pt, rotate=90] at (img.west){Design B};
    \node (designC)[left, align=center, xshift = 5pt, yshift = -28pt, rotate=90] at (img.west){Design C};
\end{tikzpicture}
}
    \vspace{-0.6cm}
    \caption{Three garment designs %
    3D garment deformation model.
}
  \label{fig:teaser}
\end{wrapfigure}
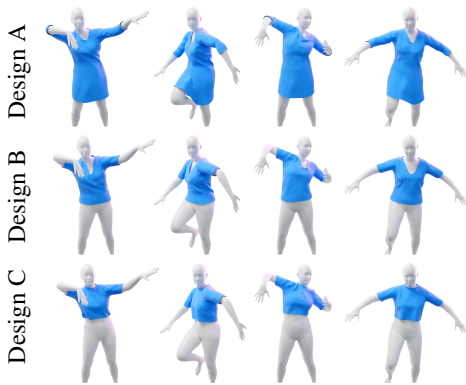

\vspace{-0.5cm}
\input{sections/introduction}

\input{sections/relatedwork}

\input{sections/representation}

\input{sections/results}

\input{sections/conclusions}

\bibliography{egbib}
\end{document}

%% file: sections/introduction.tex
\vspace{-0.2cm}
\section{Introduction}
Animating cloth is a fundamental problem in Computer Graphics and a crucial for creating 3D virtual humans that wear realistic, deformable garments.
While traditional approaches based on physics-based simulation \cite{provot1995deformation,Baraff1998,house2000cloth,selle2008robust} have achieved impressive advances in speeding up the simulation~\cite{bouaziz2014projective,tang2018gpu,narain2012arcsim,overby2017admm}, they remain computationally very expensive.

In the last few years, many learning-based methods \cite{bertiche2021pbns,patel2020tailor,vidaurre2020virtualtryon,gundogdu2019garnet,jin2020pixel,santesteban2021garmentcollisions,ma2020learning} have emerged as an alternative to physics-based solutions for clothing. 
These methods leverage modern machine learning tools, such as deep neural networks, capable of modeling highly-dimensional nonlinear functions from data, allowing them to learn how clothing deforms.   
Given such a challenging task,
designing an effective \textit{representation} to encode 3D garments is key for learning-based models to be successful. 
To this end, some methods use multilayer perceptrons (MLP) or graph-based networks to directly infer the 3D vertices position of a predefined garment~\cite{bertiche2021pbns,santesteban2021garmentcollisions,patel2020tailor,tiwari20sizer,zhang2021dynamic} or body~\cite{ma2020learning} template, but do not generalize to unseen garment mesh topologies (\textit{i.e.,} the number of vertices must be fixed).%
To circumvent this limitation, some works have explored fully-convolutional graph-based architectures \cite{vidaurre2020virtualtryon}, which make the network agnostic to mesh triangulation, but struggle with local fine-scale wrinkles.
Other approaches model 3D outfits using point clouds \cite{ma2021pop,maCVPR2021,zakharkin2021ICCV} or implicit  \cite{corona2021smplicit,tiwari21neuralgif} representations, but detailed and topologically consistent mesh outputs remain challenging. %

Instead of working in the 3D domain, some works have explored the use of 2D image-based representations to encode 3D garments, and to leverage the well-studied deep learning architectures such as GANs ~\cite{lahner2018deepwrinkles,zhang2021deep,shen2020gan} for image processing to model garment details.  
However, GANs are difficult to train (\textit{e.g.}, they easily suffer from vanishing gradient \cite{arjovsky2017towards}) and their expressiveness is limited.  

To address the limitations discussed above, we propose a novel method for learning to deform 3D garments. Our method is based on the recently introduced 2D diffusion models for image synthesis \cite{ho2020denoising}.
Our key idea is to train a \textit{conditional} diffusion model on 2D images that %
encodes 3D offsets with respect to a parametric garment template.
To this end, we first encode into a 2D layout-consistent UV texture a dataset of 3D garments simulated in different motions and body shapes.  
Then, we learn a \textit{pose-shape-and-design conditioned} diffusion model able to synthesize these 2D textures, which are converted into animations of 3D parametric garments.
Since the encoding and inference of the deformations is done in the UV space, our model is agnostic to the discretization of the garment mesh, while capable of synthesizing fine-scale wrinkles thanks to the underlying diffusion model.
Importantly, we also propose a solution to condition the model on an existing garment state, enabling the generation of temporally coherent sequences.
We qualitatively and quantitatively demonstrate that our approach is capable of generating high-quality 3D animations for a wide variety of garments, body shapes, and motions, outperforming the closest previous works for similar tasks that are based on graph neural networks or MLPs. 

%% file: sections/relatedwork.tex
\vspace{-0.2cm}
\section{Related Work}
\vspace{-0.2cm}
\noindent\textbf{Deformations in 3D space.}
Numerous methods use traditional mechanics like solving ODEs to deform cloth~\cite{nealen2006physically}. Some model mechanics at yarn level for realism~\cite{kaldor2010efficient,cirio2015efficient}, but they re computationally heavy for interactivity. Recent GPU-based solutions generate detailed outputs efficiently~\cite{wang2021gpu,wu2022gpu}.
Many works attempt to reduce the computational cost using, for example,  position-based dynamics~\cite{mueller2007pbd,kim2012long,muller2014strain}, model reduction ~\cite{de2010stable,sifakis2012fem,holden2019neuralphysics,fulton2019latent}, or adding detail to low-res simulations ~\cite{kavan2011physics,zurdo2012animating,gillette2015realtime}, but do not scale to dozens of garments. 

Physics-based methods like~\cite{umetani2011sensitive,berthouzoz2013parsing} model garment design, simulating user-defined parameters to compute 3D drape. Our approach predicts 3D drape directly from garment parameters without simulation, accommodating any design, mesh topology, body shape, or pose.
Recent techniques like differentiable cloth simulation~\cite{liang2019differentiable} and physics-based methods~\cite{um2020solver,hu2020difftaichi} show promise in optimizing parameters for desired 3D shapes. However, they're expensive, rely on simplified models, and struggle with collision handling.

To circumvent the simulation computational costs, many recent methods have been proposed to \textit{learn} 3D clothing models from data.
Guan~\etal~\cite{guan2012drape} first use 3D simulations to learn how to deform garments based on body shape and pose, but their linear model struggles with fine details.
Similarly, Xu~\etal~\cite{xu2014sensitivity} retrieve garment parts from a simulated dataset to synthesize pose-dependent clothing meshes, but do not model shape deformations. More recent methods use machine learning to predict garment deformations as a function of body pose alone ~\cite{gundogdu2019garnet}, or pose and shape~\cite{santesteban2019virtualtryon}, some even capable of learning style~\cite{patel2020tailor} or animation dynamics \cite{zhang2021dynamic} as a function of fabric parameters~\cite{wang2019intrisicspace}. However, these methods require a regressor for each garment, hindering their deployment to massive use. In contrast, our method can deform a variety of garments using the same model.

Other works emply graph neural networks to model 3D geometry ~\cite{ranjan2018generating,ma2020learning,larsen2016vaegan,gundogdu2019garnet,grigorev2023hood}. Vidaurre~\etal~\cite{vidaurre2020virtualtryon} propose a similar approach to ours using fully-convolutional graph neural networks to handle arbitrary mesh topology and parametric designs. However, their approach is limited to a single pose while we generalize to pose, shape and design. 

\vspace{0.1cm}
\noindent\textbf{Deformations in UV space.} The bijective mapping between a 3D surface and 2D plane can be done using UV maps. This representations allows to store 3D vertices in a continous space suitable to be processed with standard convolutional neural networks.
There are several methods that leverage such 2D representation to encode 3D geometry, either explicitly as input/output to the system, or implicitly, as an intermediate step followed by neural decoders \cite{zhang2022motionguided}. 
Shen~\etal~\cite{shen2020gan} introduce an image-based latent representation for sewing patterns that are decoded using a generative adversarial network. 
Su~\etal~\cite{su2022deepcloth} presents a unified neural pipeline to represent garments that can be animated and parameterized by body shape and pose. As opposed to ours, they represent the garment as a distance map with respect to the UVs of the SMPL \cite{loper2015smpl} human body mesh.
Xu~\etal~\cite{xu20193d} model the garment geometry from real photos by estimating the UV textures of a deformed garment template. 
A common strategy to modeling wrinkles consists of enhancing the coarse geometry with a network that outputs a detailed normal map~\cite{lahner2018deepwrinkles}, where the material type can be used as additional cue\cite{zhang2021deep}. 
Zhang~\etal~\cite{zhang2021dynamic} renders detailed geometry of a dynamic sequence by learning deep features attached to an initial garment template. 
In a related recent method, Korosteleva~\etal~\cite{korosteleva2022neuraltailor} follow a different paradigm to estimate the sewing pattern from a point cloud. Their method models the sewing patterns in vector space, as opposed to using UV maps. 

\vspace{0.1cm}
\noindent\textbf{Diffusion Models for Human Animation.}
Diffusion models are a type of generative model trained through the denoising diffusion process. Due to their ability to produce high-quality results~\cite{saharia2022photorealistic,ramesh2022hierarchical,croitoru2023diffusion}, outperforming GANs, they have recently received increased attention in multiple domains such as image synthesis~\cite{saharia2022palette}, 3D models~\cite{poole2022dreamfusion}, or pointclouds~\cite{luo2021diffusion}. 
Related to our goal of synthesizing garments, many recent works have explored the use of diffusion models to generate humans from text descriptions~\cite{jiang2022text2human,hong2022avatarclip,zhang2023avatarverse,kolotouros2023dreamhuman}.
These works achieve outstanding results, including photorealistic appearance and 3D details, but they do not generate pose-dependent deformations.
Furthermore, outfit and body 3D details are encoded into a single mesh or image, which precludes the explicit manipulation of the garment layer.
In contrast, our approach synthesizes pose-shape-and-design dependent deformations for explicit 3D garments, enabling the generation of layered clothed humans. 

To enable dynamic and animated downstream tasks, diffusion models have been extended to the temporal domain.
For example, they have been explored in the context of text-to-motion~\cite{hovideo,blattmann2023align,chen2023executing,singer2023makeavideo,voleti2022MCVD}, image-to-video~\cite{ni2023conditional,yu2023video}, point clouds~\cite{luo2021diffusion}, or video synthesis~\cite{luo2023videofusion,yu2023video,voleti2022MCVD}.
Diffusion models have also been used to encode human motion from sparse inputs ~\cite{du2023avatars} and from text input~\cite{tevet2022human,zhang2022motiondiffuse,kim2022flame}.
Inspired by Ho \textit{et al.} \cite{hoJMLR22cascade}, we propose to condition garment deformations on the previous garment state encoded as UV texturemap, which yields temporally-coherent 3D animation wrinkles and folds.

%% file: sections/representation.tex
\section{Overview}
Our goal is to predict 3D garment deformations from body pose, shape, and design parameters. In Section \ref{sec:garment-rep}, we introduce our garment representation: a 3D mesh encoded with an MLP network for global design, and an image-based method for folds and wrinkles from body pose and shape. Using a data-driven approach, Section~\ref{sec:data-driven-wrinkles} presents our key contribution—a diffusion-based model for predicting image-based wrinkles. Section \ref{sec:results} showcases our method's ability to animate various designs with realistic folds and wrinkles in sequences.

\section{Garment Representation}
\label{sec:garment-rep}

Our garment representation builds on top of the existing 3D parametric human models (\textit{e.g.}, \cite{loper2015smpl,joo2018total}), borrowing their shape $\upbeta$
 and pose $\uptheta$ parameterization. %
Inspired by previous works \cite{vidaurre2020virtualtryon,santesteban2019virtualtryon}, we extend SMPL body model formulation \cite{loper2015smpl} to represent a deformed garment as
\begin{equation}
	M_\text{g}(\upbeta, \uptheta, \mathbf{p}) = W(T_\text{g}(\upbeta,\uptheta, \mathbf{p}), J(\upbeta), \uptheta,\mathcal{W}), 
	\label{eq:smpl}
\end{equation}
where $W$ is a skinning function (\textit{e.g.}, linear blend skinning, or dual quaternion), $J(\upbeta) \in \mathbb{R}^{3 \times 24}$
the body joint positions, and $\mathcal{W}$ the skinning weights of a deformable garment $T_\text{g}(\cdot)$.
Our key difference with previous works \cite{vidaurre2020virtualtryon} is the representation used to encode and learn the deformable garment $T_\text{g}(\cdot)$, which allows us to learn fine-wrinkle detail while being agnostic to both the template mesh topology and the surface discretization detail.   
To this end, we propose a deformable template
\begin{equation}
	T_\text{g}(\upbeta,\uptheta,\mathbf{p}) = G_{\text{design}}(\textbf{p}) + \phi(G_{\text{wrinkles}}(\upbeta,\uptheta,\textbf{p}))
 \label{eq:garment-model}
\end{equation}
where the first term models the global deformation of the garment due to the design parameter \textbf{p}, which covers variations on the length of the garment, the length of the sleeves, and the depth of the cleavage, and the second term models the local wrinkle details due to body pose $\uptheta$, shape $\upbeta$, and design $\textbf{p}$.
In the rest of this section, we provide more details on how we model each of these terms.

The $G_{\text{design}}$ term models the global design-dependent deformations in T-pose.
In practice, we learn a function $G_{\text{design}}: |\mathbf{p}| \xrightarrow{} N_{\text{g}} \times 3 + N_{\text{g}} \times 2$ using a shallow multilayer perceptron (MLP) network that outputs  $N_{\text{g}}$ 3D vertex positions and their corresponding 2D texture coordinates of a morphable T-shirt template parameterized by sleeve length, font-and-back pannel length, and cleavage (\textit{i.e.}, the basic set of design parameters that enable the modeling of dresses, t-shirt, sweater, tops, and similar garments).
Importantly, we design our garment model such that all designs share the same UV parametrization. 

The $G_{\text{wrinkles}}$ is our key contribution to the garment model, and addresses the goal of adding pose-dependent and/or shape-dependent deformations to the output of $G_{\text{design}}$.
In contrast to previous works, which use displacements encoded in an MLP \cite{santesteban2019virtualtryon,santesteban2022snug}
or graph neural networks \cite{vidaurre2020virtualtryon},
we opt for encoding the deformations in a 2D displacement map stored as an RGB image (\textit{i.e.}, a UV texture map).
The $\phi:2 \xrightarrow{}3$ operator represents the projection operator from 2D pixel coordinates to 3D mesh coordinates which, in practice, we implement using the known mesh surface parameterization implicit in the UV coordinates.
A key design choice of our garment representation is that all $G_{\text{design}}$ outputs share a common mesh parametrization, which means that they all use \textit{the same} 2D layout despite encoding different designs.
This is a fundamental property of our representation that significantly simplifies the learning of garment wrinkles since it spatially normalizes our ground truth data.

\section{Data-Driven Diffusion-based Wrinkles}
\label{sec:data-driven-wrinkles}
\begin{figure}
\begin{minipage}{.49\textwidth}
    \includegraphics[width=\columnwidth]{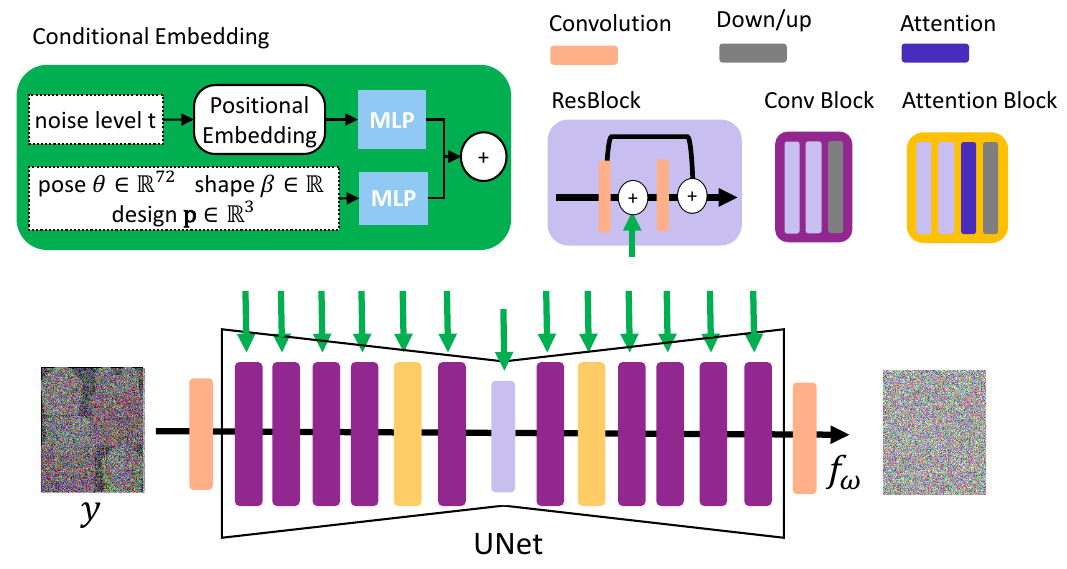}
    \caption{We use a UNet~\cite{ho2020denoising} architecture with six Resnet blocks. The conditioning vector, aggregated in the ResNet blocks, contains the pose, shape, and design parameters.
    }
    \label{fig:diffusion-model-for-garments}
\end{minipage}
\hfill
\begin{minipage}{.48\textwidth}
   \includegraphics[trim={0, 0, 0, 0pt}, clip,width=\columnwidth]{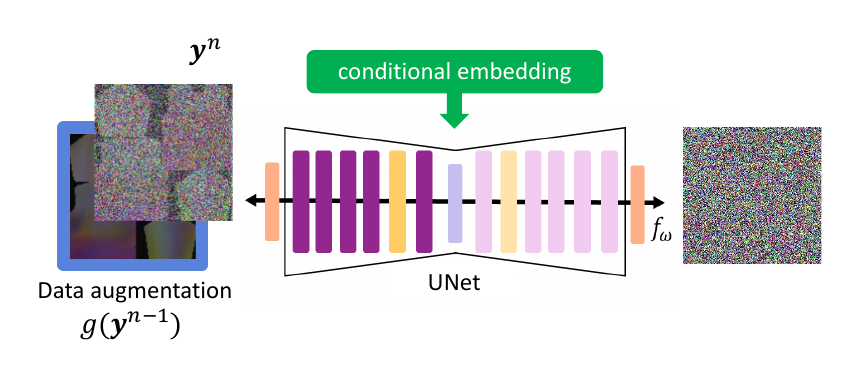}
    \caption{Temporal coherent diffusion model. To account for temporal consistency in the generated sequences while varying pose parameter, we concatenate the output of the previous frame in the sequence.}
    \label{fig:temporal-diffusion-model-for-garments}
\end{minipage}
\end{figure}

In this section, we describe how we learn the term $G_{\text{wrinkles}}$ of our garment model defined in Equation \ref{eq:garment-model} using a diffusion model.
Diffusion models~\cite{ho2020denoising} are a type of generative models that learn a target data distribution by gradually removing Gaussian noise added by a Markovian process. 
Furthermore, they can be conditioned on input variables through a conditional embedding, introduced in different layers via an MLP, as depicted in Figure~\ref{fig:diffusion-model-for-garments}. In the subsequent sections, we describe our conditional diffusion model for static scenarios (Section~\ref{sec:static}) and discuss integrating temporal constraints for generating coherent 3D garment animations (Section~\ref{sec:dynamic}).

\subsection{Pose-shape-and-designed Conditional Wrinkles} \label{sec:static}
Our goal is to learn a conditional diffusion model of the form $p(\vect{y}|\vect{c})$, where $\vect{y} \leftarrow G_{\text{wrinkles}}(\upbeta,\uptheta,\textbf{p})$ is a UVs image representing the displacement vector and $\vect{c} = \left[  \beta, \theta, \vect{p} \right] $ is the conditioning vector that includes shape $\beta$, pose $\theta$, and design $\vect{p}$ parameters. 
Given $G_{\text{wrinkles}}$, the deformation is obtained through Equation~\ref{eq:garment-model}.

Our diffusion model follows the formulation of Ho~\etal~\cite{ho2020denoising} that learns to predict the noise $\epsilon$ added at a certain step $t$ of the markovian chain. 

For training, we iteratively add random noise to the ground truth data until convergence, according to the following loss function: 
\begin{align}
	\mathcal{L}(\omega) =  \mathop{\mathbb{E}}_{ \epsilon,\vect{y}_0, t, \mathbf{c}} \left\| \, \epsilon - f_\omega \left(\mathbf{c}, \sqrt{\bar{\alpha_t}} \vect{y}_0 + \sqrt{1-\bar{\alpha_t}} \epsilon, t\right) ) \, \right\|^2_2 \, ,
\end{align}
where $f_\omega$ is the learned neural network, $\epsilon \sim \mathcal{N} (0,\mathbf{I} )$ is randomly generated Gaussian noise, $t \sim \mathcal{U} (\{1,...,T\})$ is sampled from the Uniform distribution, and $\vect{y}_0$ the ground truth sample. Finally, $\bar{\alpha_t} = \prod_{s=1}^t \alpha_s$ is the aggregated noise variance that can be computed in closed form at any timestep $t$~\cite{ho2020denoising}. %
For inference, we perform the \textit{reverse} process iteratively as: %
\begin{align}
	\vect{y}_{t-1} = \frac{1}{\sqrt{\alpha_t}} \left( \vect{y}_t - \sqrt{1 - \alpha_t} f_\omega\left(\mathbf{c},  \mathbf{y}_t, t \right)  \right)
\end{align}
At the beginning of the diffusion process ($t=\textrm{T}$) the initial value for $\vect{y}_\textrm{t=T}$ is virtually indistiguishable from Gaussian noise. Then, iteratively, from $t=\textrm{T}$ until $t=1$ this image is denoised by substracting the outputs predicted by the neural network $f_\omega$ until we obtain an approximation of $\vect{y}_0$.

\subsection{Temporally Coherent Garment Wrinkles}
\label{sec:dynamic}
Using the diffusion model described in Section \ref{sec:static} we can generate plausible wrinkles conditioned on pose, shape, and design.
However, if we sample the model for a sequence of poses, we will obtain a non-temporally coherent animation: consecutive frames will exhibit significantly different deformations.
This is due to the generative nature of the model, since we sample it with random noise it can produce \textit{different} results even for the same condition.
This prevents the conditional model from Section \ref{sec:static} to generate temporally coherent animations of garments. 

To tackle this issue, we take inspiration from cascade models for high-resolution image synthesis that condition a sample on a low-resolution version of the target image to drive the diffusion process towards a specific target \cite{hoJMLR22cascade}.
We propose to use a similar strategy to enforce temporal coherency. 
To this end, to synthesize the garment deformations at frame $n$, we further condition our diffusion model from Section \ref{sec:static} on the output image $\vect{y}^{n-1}$ of the previous frame $n-1$ of the sequence. 
In practice, we implement this by adding into our neural network $f_\omega$ an extra input $g(\vect{y}^{n-1})$ that is concatenated to $\mathbf{y}$.
To avoid overfitting this new conditional signal to the training ground truth values of $\vect{y}^{n-1}$, we apply several perturbations $g()$ to the UV images that will be described in the implementation section. Figure \ref{fig:temporal-diffusion-model-for-garments} illustrates this process.

%% file: sections/results.tex
\section{Results and Evaluation}
\label{sec:results}

\begin{wrapfigure}{r}{0.5\linewidth}
\centering
\vspace{-1.2cm}
\resizebox{\linewidth}{!}{%
\begin{tikzpicture}
    \node (img) at (0,0) {\includegraphics[trim={0 0 340 0},clip,width=\linewidth]{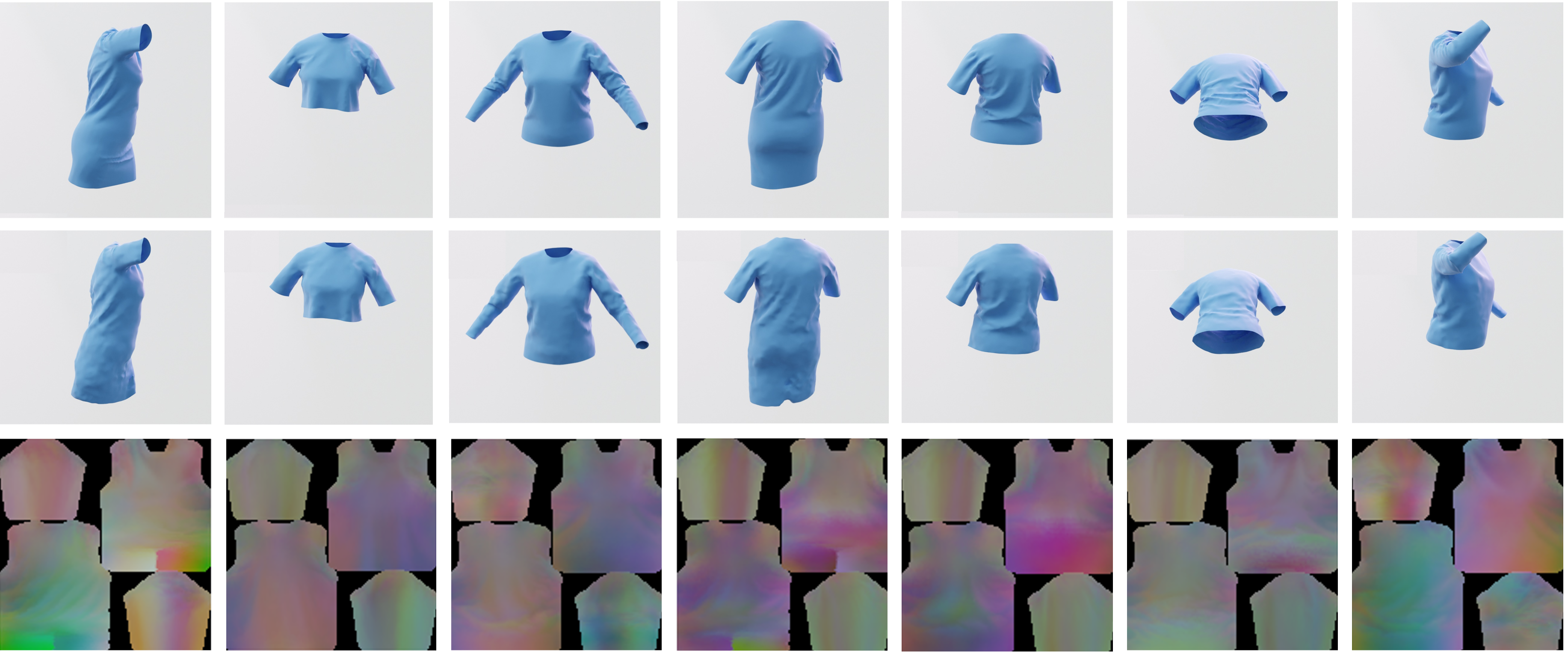}};
    \node (input)[left, align=center, xshift = -10pt, yshift = 75pt, rotate=90] at (img.west){\scriptsize{Ground truth}\\\scriptsize{simulation}};
    \node (input)[left, align=center, xshift = -10pt, yshift = 25pt, rotate=90] at (img.west){\scriptsize{Reconstruction}\\\scriptsize{from UV image}};
    \node (input)[left, align=center, xshift = -10pt, yshift = -25pt, rotate=90] at (img.west){\scriptsize{UV image (\textit{i.e.}, } \\\scriptsize{our train data)}};
\end{tikzpicture}
}
\caption{Dataset samples. We simulate garments on different bodies (top), which we convert into a UV image (bottom) that faithfully reconstruct the original garment (middle).}
\label{fig:dataset}
\end{wrapfigure}
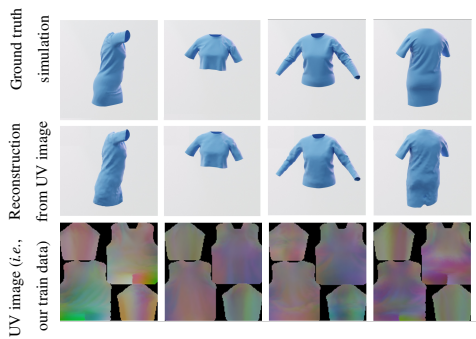 
\subsection{Dataset}
\label{sec:dataset}
To train our method, we first build a large dataset of UV-encoded deformations for a variety of garment designs worn by different body poses and shapes.
To this end, we first manually create a deformable template of a 3D garment parameterized by length, sleeve, and cleavage.
Importantly, all designs sampled by this parametric template share the same 3D-to-2D parameterization (\textit{i.e.}, the same UV layout).

Using a state-of-the-art cloth simulator~\cite{narain2012arcsim}, we statically simulate a wide variety of garment designs worn by different SMPL \cite{loper2015smpl} body sequences from AMASS dataset \cite{AMASS:ICCV:2019}.
For each simulated frame, similar to \cite{santesteban2019virtualtryon}, we project the deformed garment  into a canonical state (\textit{i.e.}, T-pose)
by unposing the mesh using the inverse transform of the skinning weights of the underlying body pose.
We then compute the per-vertex offset between the unposed mesh and the template mesh and store it as an RGB value of a texture image using the known 3D-to-2D mapping. 
Following this strategy and using standard barycentric coordinates, we can assign a value to all pixels of the texture map. 
Generated texture maps effectively encode the 3D garment deformations in a convenient 2D image format that can be exploited with a diffusion model.
Following the reverse process, we can reconstruct a deformed 3D garment by querying the UV texture value of each vertex, and then posing the garment using the skinning values of the target pose, as shown in Equation~\ref{eq:smpl}.

Figure~\ref{fig:dataset} depicts a few samples of our dataset including ground truth simulations (top), the corresponding UV texture encoding deformations (bottom), and the reconstructed 3D garment from the UV images (middle).
In practice, we simulate 17 designs of garments (7 different garment lengths, 6 different sleeves, and 4 types of cleavage) in 52 sequences, and generate a $128\times128$ pixels UV textures to encode the deformation of each frame.
We train on 11 designs and leave out 6 designs and 5 sequences for validation.
Once trained, our model generalizes to unseen combinations of garment parameters, producing plausible deformations for new garments.

\subsection{Network Architecture and Implementation Details}
\label{sec:implementation-details2}
Our neural network $f_\omega$ from Section \ref{sec:static} is implemented as a symmetrical UNet that consists of six downsampling residual layers. The fifth layer includes a spatial self-attention block, which has been proven successful in performing global reasoning~\cite{vaswani2017attention}. Each ResNet layer has two layers, and the number of output channels for each UNet block is 128, 128, 256, 256, 512, 512.
The conditional embedding is implemented as a 2-layer MLP with a 128-feature vector.
We train our model using a batch size of 8 for 100 diffusion steps using a single NVidia Titan X.
Our unoptimized diffusion model can be plugged into recent works on diffusion distillation techniques  \cite{sauer2023adversarial,meng2023distillation} to achieve interactive frame rates.

Our temporally coherent diffusion model architecture described in Section \ref{sec:dynamic} is analogous to the design of $f_\omega$ described above.
The key difference is the input, which is expanded with the previous frame of the sequence. 
The architecture does not need to be updated as both images are concatenated, only changing the depth of the intermediate outputs. Because at this step the previous frame will already be converged, it will be a strong signal for the network and potential cause of overfitting. To avoid it, we apply a data augmentation process consisting of randomly applying Gaussian blur and color jitter effects. 
Since we utilize previous frames from the ground truth during training, they can act as a strong signal for the network and potentially lead to overfitting. To avoid it, we implement a data augmentation strategy during training, which consists of randomly applying Gaussian blur and color jitter effects to the previous frame. Furthermore, to prevent the network's dependency on the previous frame, we randomly erase portions of the image during training.

\subsection{Quantitative Evaluation.} 
\begin{figure}[t!]
     \centering
     \includegraphics[width=\linewidth]{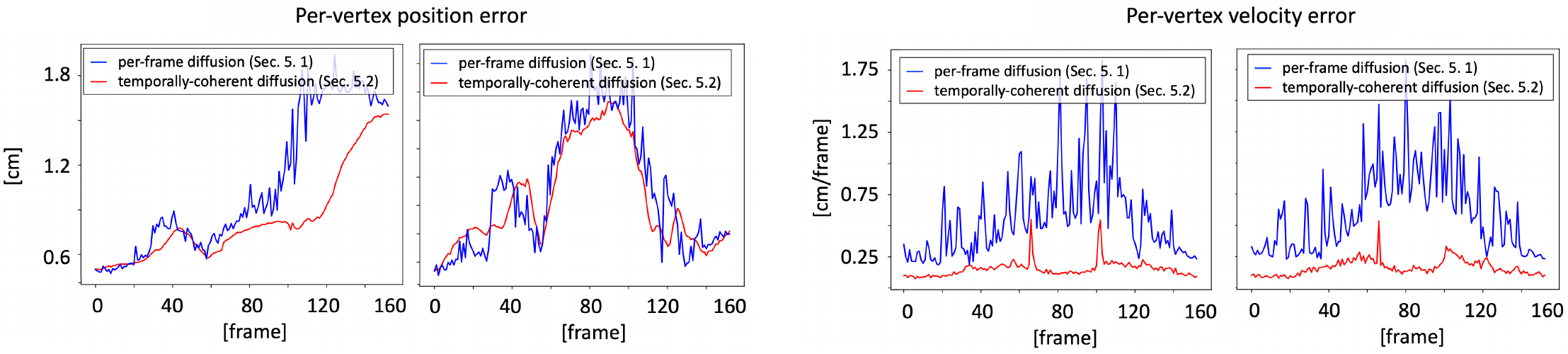}
     \caption{Quantitative evaluation of our temporally-coherent diffusion model (in red) and per-frame diffusion model (in blue), in two test sequences.
     Since our temporal model is conditioned on the previous deformation state of the garment, the resulting animations are temporally smooth and closer to the ground truth surface.
     }
     \label{fig:temporally-coherent-evaluation}
     \vspace{-0.5cm}
\end{figure}

Figure \ref{fig:temporally-coherent-evaluation} presents a quantitative evaluation of our proposed diffusion model. The blue curve represents the model conditioned on pose-shape-and-design (Section \ref{sec:static}), while the red curve represents our temporally-coherent model additionally conditioned on the previous state of the garment (Section \ref{sec:dynamic}).
For each model, we plot the per-vertex position error (left) and the velocity error (right) compared to two ground truth simulations on two validation garments designs (top and bottom) unseen at train time. 

Our temporally-coherent diffusion model consistently outperforms the static model only conditioned on pose-shape-and-design, delivering lower and much more stable per-frame vertex error.
This is clearly observed at the vertex velocity error plots (Figure \ref{fig:temporally-coherent-evaluation}, right). Our temporal model (in red), conditioned on the previous garment state, closely matches the ground truth velocity, while a static per-frame deformation synthesis (in blue) significantly and incoherently differs from the ground truth. 
A qualitative visualization of this plot can be found in the supplementary video, showcasing smooth surface deformations over time when using our temporal model.

\vspace{-0.2cm}
\subsection{Qualitative Evaluation.} 
\begin{figure}[t!]
\begin{minipage}[t]{0.46\textwidth}
    \centering
    \begin{tikzpicture}
    \node (img) at (0,0) {\includegraphics[width=0.93\linewidth]{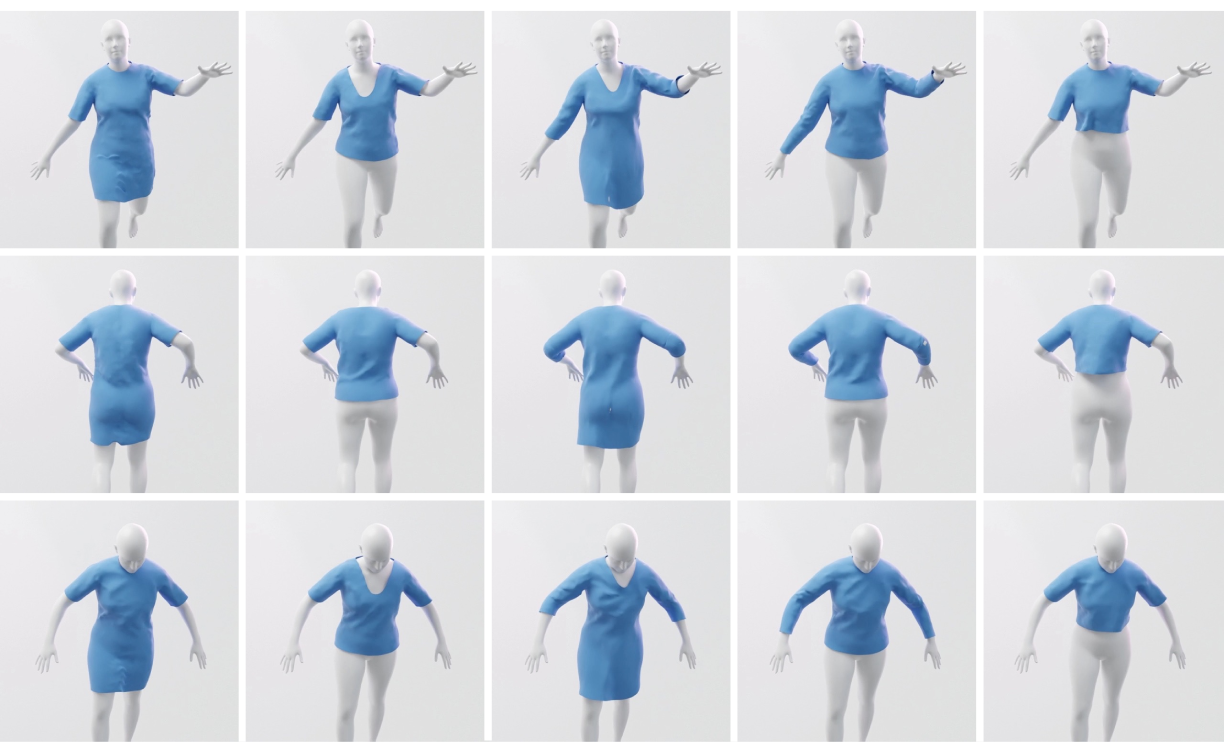}};
    \end{tikzpicture}
    \caption{Qualitative results of five test garment designs (columns) deformed using our diffusion-based model driven by a test motion from AMASS dataset. 
    }
    \label{fig:predictions-mean-shape}
\end{minipage}
\hfill
\begin{minipage}[t]{0.515\textwidth}
\centering
    \begin{tikzpicture}
    \node (img) at (0,0) {\includegraphics[trim={0 0 460 0}, clip,width=\linewidth]{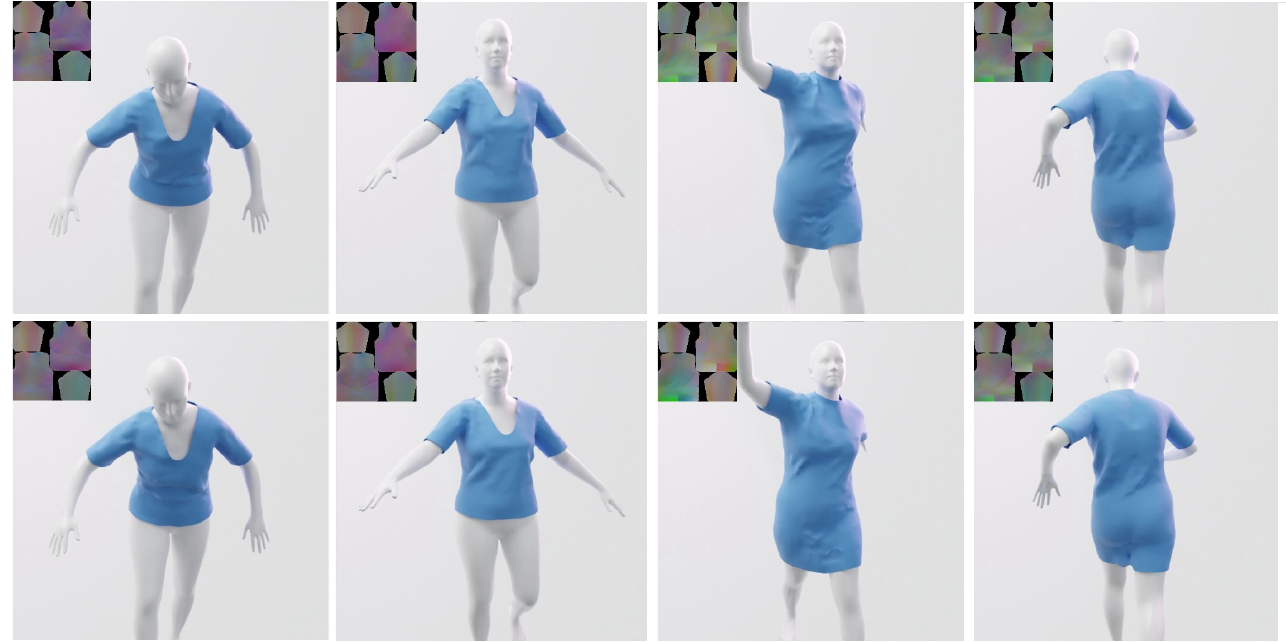}};
     \end{tikzpicture}
    \caption{Ground truth (top) vs our model (bottom) on test sequences. Our predictions closely match the folds and wrinkles obtained with physics-based methods.}
    \label{reconstruction-estimation}
\end{minipage}
\end{figure}

Figure \ref{reconstruction-estimation} compares our diffusion model results (bottom) with ground truth simulations (top) across diverse garment designs and body poses unseen during training. Despite complex dynamic deformations and diverse folds, our model accurately matches the ground truth. Supplementary video provides more animated results.

Figure \ref{fig:predictions-mean-shape} showcases five validation garment designs worn by differently posed bodies, exhibiting rich folds and wrinkles that match body pose realistically. This mosaic highlights the expressive power of our diffusion-based model. Please refer to the supplementary video for animated results. Similarly, Figure \ref{fig:teaser} displays natural 3D clothing deformations across pose, shape, and design during a hip-hop dancing motion from the AMASS dataset.

\vspace{0.2cm}
\subsection{Comparison to state-of-the-art.}
\begin{figure}[t!]
\begin{minipage}[t]{0.48\textwidth}

     \centering
     \begin{tikzpicture}
    
     \node[] (img) at (0,0)
     {\includegraphics[trim = {0 0 0 140pt}, clip,width=\linewidth]{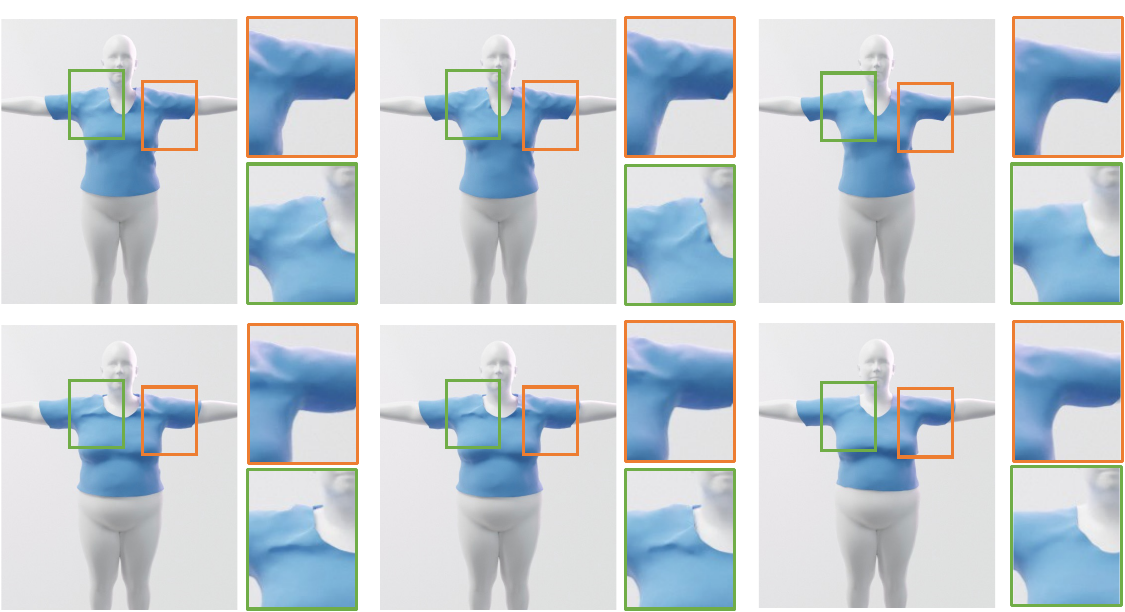}};
    \node (a-label) [align=left, xshift = -52pt, yshift = 3 pt] at (img.north)
    {\scriptsize{Ground truth}};
     \node (a-label) [align=left, xshift = -5pt, yshift = 3 pt] at (img.north)
    {\scriptsize{Ours}};
         \node (a-label) [align=left, xshift = 50pt, yshift = 3 pt] at (img.north)
    {\scriptsize{Vidaurre \etal ~{\cite{vidaurre2020virtualtryon}}}};
     \end{tikzpicture}
     \caption{Qualitative comparison with \cite{vidaurre2020virtualtryon} for a garment design unseen at train time. Our diffusion model predicts 3D deformations that closely match the ground truth, while the state-of-the-art method \cite{vidaurre2020virtualtryon} produces oversmooth deformations.}
     \label{fig:sca-comparison}

\end{minipage}%
\hfill
\begin{minipage}[t]{0.48\textwidth}

     \centering
     \begin{tikzpicture}
    
     \node[] (img) at (0,0)
     {\includegraphics[trim = {0 0 0 1200pt}, clip, width=\linewidth]{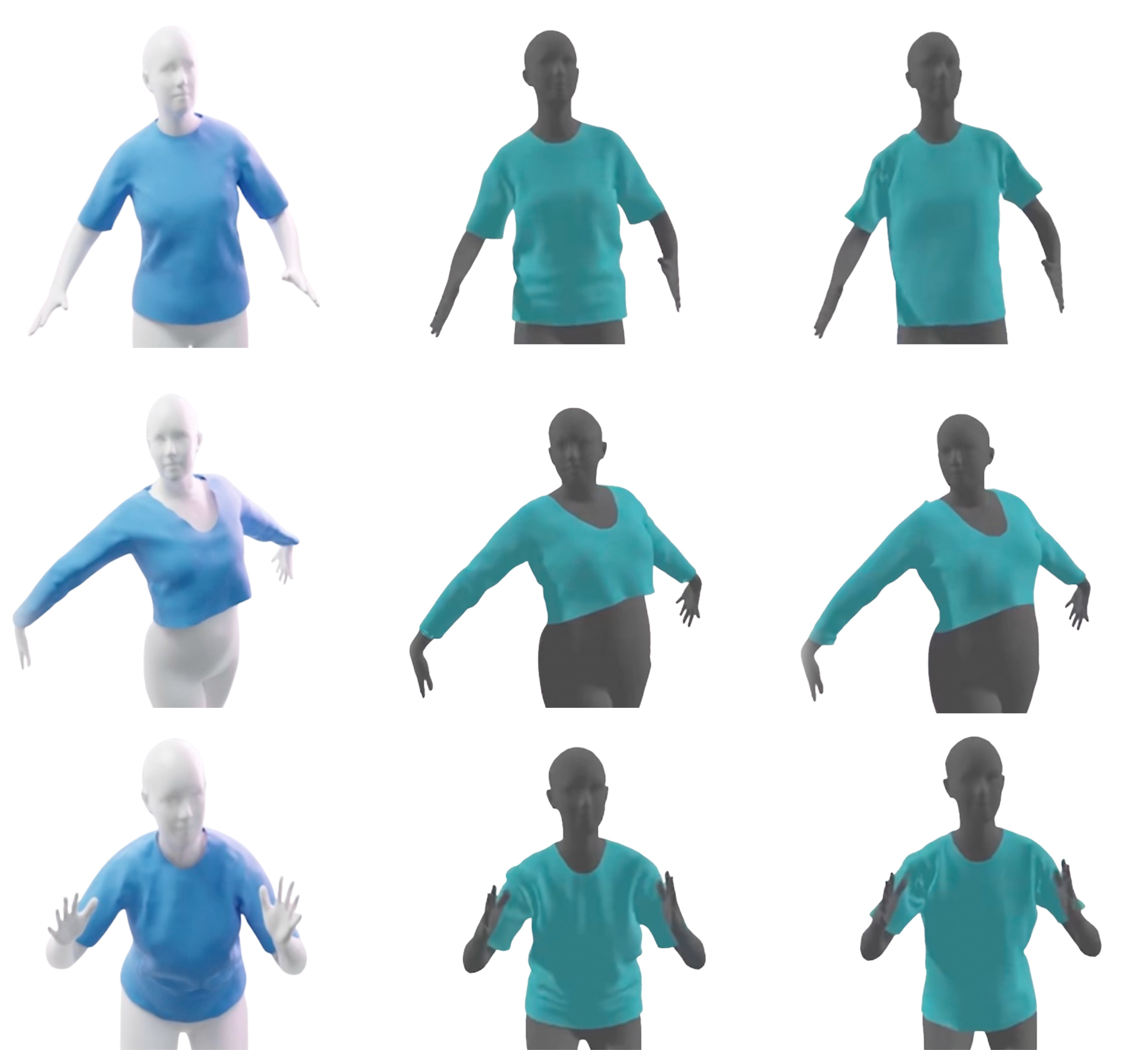}};
    
    \node (a-label) [align=left, xshift = -52pt, yshift = 3 pt] at (img.north)
    {\scriptsize{Ours}};
     \node (a-label) [align=left, xshift = -5pt, yshift = 3 pt] at (img.north)
    {\scriptsize{SNUG \cite{santesteban2022snug}}};
         \node (a-label) [align=left, xshift = 50pt, yshift = 3 pt] at (img.north)
    {\scriptsize{HOOD \cite{grigorev2023hood}}};
     \end{tikzpicture}
     \caption{Qualitative comparison with SNUG~\cite{santesteban2022snug} and HOOD \cite{grigorev2023hood}. Our method is on par with their quality, but offers different features. SNUG is limited to a single garment, and HOOD does not model explicit garment design parameters}
     \label{fig:sca-comparison2}

\end{minipage}%
 \end{figure}

Figure \ref{fig:sca-comparison} presents a qualitative comparison with the method of Vidaurre \textit{et al.} \cite{vidaurre2020virtualtryon}.
We show garment deformations obtained by each method for a test design in various body shapes.
Notice that \cite{vidaurre2020virtualtryon} does not model pose-dependent deformations, hence we limit our comparison to T-pose avatars. 
Our method obtains deformations that closely match the ground truth simulation, which demonstrates that our diffusion-based model is more expressive than the fully-convolutional graph model of \cite{vidaurre2020virtualtryon}.

In Figure \ref{fig:sca-comparison2} and the supplementary video, we present a qualitative comparison of our method with self-supervised methods SNUG \cite{santesteban2022snug} and HOOD \cite{grigorev2023hood}. Quantitative comparison is challenging due to significant differences in representations, models, and objectives. For instance, SNUG models dynamics but is limited to a single garment, while HOOD produces compelling results for unseen garments but lacks generativity and explicit incorporation of design parameters. Despite these disparities, our method demonstrates comparable deformations to state-of-the-art methods, using a compact image-based representation.

%% file: sections/conclusions.tex
\section{Conclusions}
We presented DiffusedWrinkles, a method for synthesizing 3D garment deformations based on pose, shape, and design using a 2D diffusion-based model. Our approach enables representation of a wide range of 3D garment designs with consistent 2D layouts, allowing for image-based diffusion models. We achieve temporally-coherent 3D deformations in animations through cascade architectures, resulting in compelling results depicting body shape, pose, and designs.

Despite the step forward of our method in data-driven garment techniques, it has limitations. Similar to \cite{santesteban2019virtualtryon,patel2020tailor}, body-garment collisions remain a challenge, addressed at inference by pushing problematic vertices outward. Moreover, the expressivity of the diffusion model limits generalization; as training samples increase, results may become overly smooth. This could be improved by employing a Latent Diffusion Model for more expressive image subspaces.
Finally, dynamic effects are currently not modeled. Our approach takes as input the current state of the garment which yields a temporally-coherent output, but a longer temporal window and a more complex architecture are needed to model time-dependent effects.